%% file: main.tex
\documentclass[acmtog]{acmart}
\acmSubmissionID{1209} 

\input{header}

\author{Chuhao Chen}
\affiliation{
  \institution{University of California San Diego}
  \country{USA}
}
\email{chc091@ucsd.edu}

\author{Isabella Liu}
\affiliation{
  \institution{University of California San Diego}
  \country{USA}
}
\email{lal005@ucsd.edu}

\author{Xinyue Wei}
\affiliation{
  \institution{University of California San Diego}
  \country{USA}
}
\affiliation{
  \institution{Hillbot Inc.}
  \country{USA}
}
\email{xiwei@ucsd.edu}

\author{Hao Su}
\affiliation{
  \institution{University of California San Diego}
  \country{USA}
}
\affiliation{
  \institution{Hillbot Inc.}
  \country{USA}
}
\email{haosu@ucsd.edu}

\author{Minghua Liu}
\affiliation{
  \institution{Hillbot Inc.}
  \country{USA}
}
\email{m@hillbot.ai}

\copyrightyear{2025}
\acmYear{2025}
\setcopyright{acmlicensed}\acmConference[SA Conference Papers '25]{SIGGRAPH Asia 2025 Conference Papers}{December 15--18, 2025}{Hong Kong, Hong Kong}
\acmBooktitle{SIGGRAPH Asia 2025 Conference Papers (SA Conference Papers '25), December 15--18, 2025, Hong Kong, Hong Kong}
\acmDOI{10.1145/3757377.3763845}
\acmISBN{979-8-4007-2137-3/2025/12}

\begin{document}

\title{FreeArt3D: Training-Free Articulated Object Generation using 3D Diffusion}

\input{sections/abstract}

\begin{teaserfigure}
    \includegraphics[width=\textwidth]{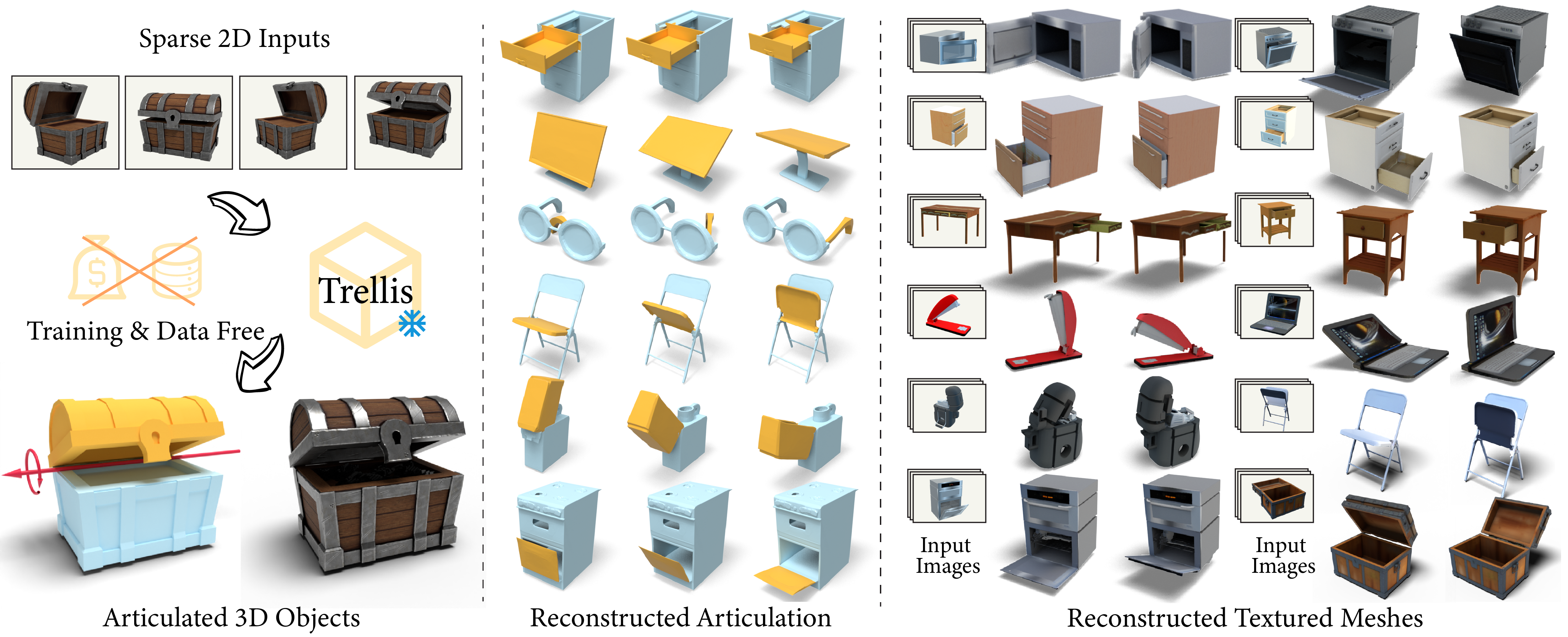}
    \vspace{-2em}
    \caption{Facing the scarcity of 3D articulated object datasets, we propose FreeArt3D, a novel training-free pipeline for generating 3D articulated objects without requiring any specific dataset. We employ a per-shape optimization strategy by repurposing a pretrained 3D diffusion model—originally developed for static object generation—as a 3D guidance prior. This enables the reconstruction of high-quality 3D articulated objects from sparse input views within minutes. Unlike recent methods that rely on part retrieval or template meshes and often miss fine details and overlook texture, our approach strictly follows the input prompts and produces high-quality results with faithful geometry, realistic textures, and accurate articulation structures.
    } 
    \label{fig:teaser}
\end{teaserfigure}
\maketitle

\input{sections/introduction}

\input{sections/related_work}
\input{sections/method}
\input{sections/experiment}

\input{sections/conclusion}

\clearpage

\bibliographystyle{ACM-Reference-Format}
\bibliography{bib}

\end{document}

%% file: header.tex
\usepackage{multirow} 

%% file: sections/abstract.tex
\begin{abstract}
Articulated 3D objects are central to many applications in robotics, AR/VR, and animation. Recent approaches to modeling such objects either rely on optimization-based reconstruction pipelines that require dense-view supervision or on feed-forward generative models that produce coarse geometric approximations and often overlook surface texture. In contrast, open-world 3D generation of static objects has achieved remarkable success, especially with the advent of native 3D diffusion models such as Trellis. However, extending these methods to articulated objects by training native 3D diffusion models poses significant challenges. In this work, we present FreeArt3D, a training-free framework for articulated 3D object generation. Instead of training a new model on limited articulated data, FreeArt3D repurposes a pre-trained static 3D diffusion model (e.g., Trellis) as a powerful shape prior. It extends Score Distillation Sampling (SDS) into the 3D-to-4D domain by treating articulation as an additional generative dimension. Given a few images captured in different articulation states, FreeArt3D jointly optimizes the object’s geometry, texture, and articulation parameters—without requiring task-specific training or access to large-scale articulated datasets. Our method generates high-fidelity geometry and textures, accurately predicts underlying kinematic structures, and generalizes well across diverse object categories. Despite following a per-instance optimization paradigm, FreeArt3D completes in minutes and significantly outperforms prior state-of-the-art approaches in both quality and versatility. Please check
our website for more details: \color{blue}{\url{https://czzzzh.github.io/FreeArt3D}}.
\end{abstract}

%% file: sections/introduction.tex
\section{Introduction}

Articulated 3D objects—ranging from everyday items like glasses and phones to larger assets such as appliances and furniture—are ubiquitous in our surroundings. Accurately capturing their geometry, appearance, and underlying kinematic structures is crucial for a wide range of applications, including robotics, digital twins, AR/VR, and animation pipelines.

Building on this need, recent efforts in articulated 3D generation and reconstruction have followed two main directions. One line of work~\cite{tseng2022cla,wei2022self,liu2023paris} focuses on optimization-based methods—often leveraging NeRFs or 3D Gaussian Splatting—to recover fine-grained articulated geometry. While these techniques can produce decent results, they are computationally intensive, require dense multi-view supervision, limiting their scalability. In parallel, several feed-forward models~\cite{le2024articulate,liu2024singapo,chen2024urdformer} have been proposed to directly reconstruct or generate articulated objects from more accessible inputs such as object categories, articulation graphs, or even single-view images and text prompts. Although faster, many of these models rely on coarse geometric approximations using bounding boxes, predefined templates, or small part retrieval databases. As a result, the reconstructed shapes often lack structural realism and fine detail, and many methods fail to model surface texture, further diminishing visual fidelity.

In contrast, open-world 3D generation of static objects has seen greater success. Early efforts leveraged 2D priors from pre-trained diffusion models—either through score distillation techniques~\cite{poole2022dreamfusion} or by fine-tuning to enable multi-view synthesis~\cite{liu2023zero}—followed by the training of sparse-view reconstruction models~\cite{liu2023one}. More recently, several open-world approaches~\cite{xiang2024structured,zhang2024clay} have emerged that train native 3D diffusion models directly in the 3D domain. Notably, models like Trellis~\cite{xiang2024structured} have demonstrated that by adopting effective 3D representations, carefully designing 3D architectures and algorithms, and scaling up training data, it is possible to generate high-quality 3D geometry with realistic textures, while maintaining strong generalization across diverse object categories.

Extending the success of static object generation to articulated 3D objects remains largely underexplored. A natural approach is to train a native 3D diffusion model tailored for articulated objects, but two key challenges arise. First, articulated objects require a representation that jointly captures multi-part geometry and kinematic structure. Second, existing datasets~\cite{xiang2020sapien} are limited—typically containing only a few thousand shapes—several orders of magnitude smaller than those for static objects, making it difficult to train diffusion models effectively.

To address these challenges, we introduce FreeArt3D, a training-free framework for articulated 3D object generation from sparse input views. Rather than training a new model from scratch, FreeArt3D leverages a pre-trained 3D diffusion model—such as Trellis—originally trained on static objects, as a general 3D shape prior. It extends the concept of Score Distillation Sampling (SDS), which was originally developed for guiding 3D generation using 2D diffusion models, into the 3D-to-4D domain by treating articulation as an additional generative dimension. 

Given a few images of an object captured at different articulation states, FreeArt3D optimizes an articulated object representation from scratch. This representation includes separate geometries for the static body and the movable part, joint-related parameters (e.g., rotation axis and pivot point), and the joint state corresponding to each input image. By transforming the component geometries according to the estimated joint parameters and per-image joint configurations, we synthesize the 3D object in each articulation state. These synthesized 3D models, paired with their corresponding input images, are then fed into the 3D diffusion model to compute guidance signals that drive the optimization of geometry and joint representations. To ensure robust convergence and high-quality textured mesh, we introduce carefully designed strategies for shape normalization, initialization, and post-processing. 

FreeArt3D enables high-fidelity articulated object generation without requiring task-specific training or access to large-scale articulated datasets. We compare FreeArt3D with recent state-of-the-art models and show that our method significantly outperforms them by a large margin in terms of geometry, appearance, and kinematic structure metrics. FreeArt3D not only generates higher-quality geometry and textures that better adhere to user input but also predicts more accurate articulated structures. Moreover, unlike previous methods that are limited to a small set of predefined categories, our approach demonstrates stronger generalizability and performs well across a wide range of object types, as shown in Figure~\ref{fig:teaser}. Although our method adopts a per-shape optimization paradigm, it completes within a reasonable time—typically just a few minutes. We also demonstrate that our method can be easily extended to support multiple joints and achieves robust performance on real-world captured images.

%% file: sections/related_work.tex
\section{Related Work}
\subsection{Understanding, Reconstruction, and Generation of 3D Articulated Objects}

Prior research has extensively investigated the understanding of articulated objects, focusing on tasks such as detecting movable parts from a single image~\cite{sun2023opdmulti,jiang2022opd} or from video sequences~\cite{qian2022understanding}. A substantial body of work targets joint parameter estimation—such as joint axes, pivot points, and articulation angles—given various forms of input, including point clouds~\cite{fu2024capt,liu2023category,xu2022unsupervised,li2020category,yan2020rpm,jiang2022ditto,wang2019shape2motion}, RGB-D images~\cite{jain2022distributional,liu2022toward,che2024op,jain2021screwnet,abbatematteo2019learning,hu2017learning}, video~\cite{liu2020nothing}, or 4D dynamic point clouds~\cite{liu2023building}. Beyond static analysis, active perception methods have been proposed to facilitate more accurate estimation~\cite{zeng2024mars,yan2023interaction}. Recently, vision-language models (VLMs) have also been finetuned for joint estimation tasks~\cite{huang2024a3vlm}. Several methods address temporal modeling of articulated objects, such as tracking and pose estimation over time~\cite{heppert2022category,weng2021captra}.

Beyond understanding, numerous methods aim to reconstruct or generate articulated objects. A popular approach is per-instance optimization using Neural Radiance Fields (NeRF)~\cite{tseng2022cla,wu2022d,liu2023paris,weng2024neural,mu2021sdf,song2024reacto,wang2024sm,deng2024articulate,swaminathan2024leia} or 3D Gaussian Splatting~\cite{wu2025reartgs}. While effective in capturing object-specific geometry and appearance, they are typically slow to optimize and rely on densely sampled, posed images across multiple articulation states—conditions that are challenging to satisfy in real-world scenarios, thus limiting their scalability.

To improve scalability, recent works explore feedforward approaches for articulated object reconstruction and generation from object categories, articulation graphs~\cite{liu2024cage}, single images~\cite{liu2024singapo,chen2024urdformer,dai2024automated,kawana2023detection}, multi-view images~\cite{mandi2024real2code,heppert2023carto,gadi2023rosi,zhang2021strobenet}, text~\cite{su2024artformer}, or static 3D mesh~\cite{qiu2025articulate}. These models leverage diffusion models~\cite{gao2024meshart,lei2023nap,luo2024physpart}, vision-language models~\cite{le2024articulate}, or large-scale reconstruction frameworks~\cite{gao2025partrm}. However, they often simplify object geometry using coarse approximations such as bounding boxes, template parts, or retrieval from small databases, which limits their ability to capture fine-grained and realistic shape details. Moreover, most of these methods focus solely on geometry, neglecting the reconstruction of accurate textures, which reduces the visual fidelity and realism of the results.

\vspace{-0.3em}
\subsection{Optimization-Based 3D and 4D Generation}
\vspace{-0.3em}
Early open-world 3D generation methods rely on per-shape optimization, iteratively refining 3D representations to match input text or images using supervision from pre-trained 2D models (e.g., CLIP~\cite{radford2021learning}, Stable Diffusion~\cite{rombach2022high}) via differentiable rendering. DreamFusion~\cite{poole2022dreamfusion} pioneered this direction with Score Distillation Sampling (SDS), which distills denoising gradients from diffusion models. Despite their flexibility, these methods~\cite{lin2023magic3d,chen2023fantasia3d,wang2023prolificdreamer,wang2023score,jain2022zero,sanghi2022clip} often suffer from multi-view inconsistency (e.g., the Janus problem) and slow convergence.

Recent works extend SDS-based optimization to 4D generation~\cite{bahmani20244d,li2024dreammesh4d,zhang20244diffusion,singer2023text,ren2023dreamgaussian4d,sun2024eg4d,zhao2023animate124,jiang2023consistent4d}, enabling dynamic scene synthesis by jointly optimizing spatial and motion parameters (e.g., deformation fields) without explicit motion supervision. These methods leverage text-to-image or text-to-video diffusion models and apply losses across time steps or camera views to ensure both appearance and motion consistency.

While prior work focuses on general dynamic objects, such as character animations, we target articulated object reconstruction, which involves rigid part-based motion and requires disentangled geometry and articulation. Instead of relying on 2D or video diffusion models—often prone to 3D inconsistencies—we utilize static 3D diffusion priors~\cite{xiang2024structured} to guide full-shape generation, significantly improving consistency across views and poses.

\vspace{-0.3em}
\subsection{Feed-Forward 3D Static Object Generation}
\vspace{-0.3em}
Feed-forward 3D generation methods have recently gained prominence, offering significant improvements in speed and quality by learning from large 3D datasets~\cite{deitke2023objaverse,deitke2023objaversexl} to directly synthesize 3D representations without iterative optimization. These methods generally fall into two main paradigms. The 2D-lifting-to-3D paradigm either trains models to reconstruct 3D shapes directly from single image inputs~\cite{hong2023lrm,zou2024triplane,tochilkin2024triposr} or adopts a two-stage approach: first generating multi-view images using 2D priors~\cite{liu2023zero,shi2023zero123++,liu2023syncdreamer,shi2023mvdream,long2024wonder3d}, followed by feed-forward 3D reconstruction from sparse views~\cite{li2023instant3d,liu2023one,liu2024one,liu2024meshformer,wei2024meshlrm,wang2024crm,tang2024lgm,xu2024instantmesh,wu2024unique3d}. More recently, the 3D-native generation paradigm—pioneered by models~\cite{wu2024direct3d} such as CLAY~\cite{zhang2024clay}, Trellis~\cite{xiang2024structured}, and Hunyuan3D~\cite{zhao2025hunyuan3d}—directly models 3D geometry using generative techniques, often through a two-step process of geometry synthesis followed by texture generation. Enabled by large-scale datasets and advances in 3D representations, VAEs, diffusion models, and autoregressive methods~\cite{siddiqui2024meshgpt,chen2024meshanything}, this paradigm achieves more accurate geometry and higher-quality textures.

\begin{figure*}
    \centering
    \includegraphics[width=\linewidth]{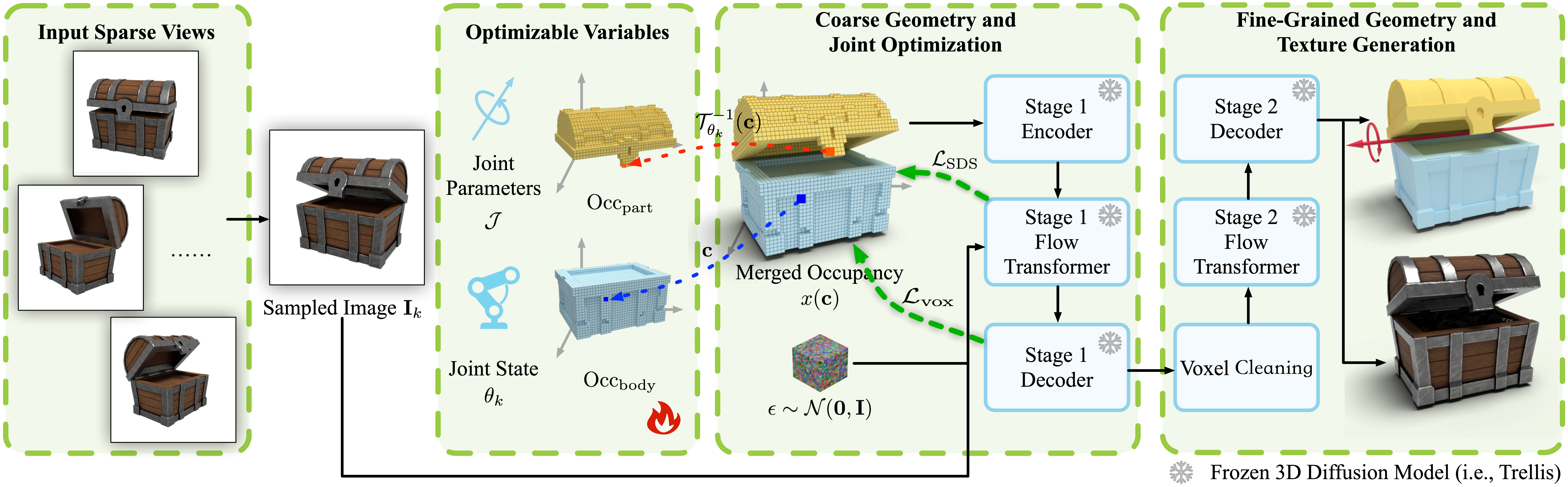}
    \vspace{-2em}
    \caption{FreeArt3D employs a per-shape optimization strategy. Given sparse-view images of different joint states as input, we jointly optimize two separate geometries—one for the body and one for the movable part—the joint parameters $\mathcal{J}$ (e.g., joint axis, pivot point), and optionally the joint states $\theta_k$, if they are not provided. During the coarse geometry and joint optimization stage, we sample an image $\mathbf{I}_k$ corresponding to joint state $\theta_k$ at each iteration and aim to construct an occupancy grid of the full object under this configuration. To achieve this, we query two hash grids and transform the coordinates according to the current joint parameters $\mathcal{J}$ and joint state $\theta_k$. The merged occupancy grid is then passed to a pretrained 3D diffusion model, Trellis~\cite{xiang2024structured}, along with the image $\mathbf{I}_k$ to provide gradient guidance for optimization. After completing the coarse stage, we clean the merged, optimized voxels and input them into the pretrained second-stage diffusion and VAE models to generate fine-grained geometry and realistic textures.
}
\vspace{-0.7em}
    \label{fig:pipeline}
\end{figure*}

%% file: sections/method.tex
\section{Method}

\subsection{Overview}
\label{sec:overview}

We propose \emph{FreeArt3D}, a training-free framework for articulated 3D object generation that formulates the problem as an optimization task per object instance, as shown in Figure~\ref{fig:pipeline}. Instead of training a feedforward model, we leverage the prior of a pre-trained 3D diffusion model (e.g., Trellis ~\cite{xiang2024structured}), originally developed for static object generation, to supervise the optimization of articulated shapes. Given a sparse set of $K$ input RGB images $\{\mathbf{I}_k\}_{k=1}^K$ of an articulated object captured under varying articulation states $\{\theta_k\}_{k=1}^K$, our goal is to reconstruct high-fidelity textured meshes of the object along with its underlying joint structure.

For simplicity, we consider objects with a single joint, and decompose the object into two components: a static body with textured mesh $\mathcal{M}_\text{body}$, and a movable part with textured mesh $\mathcal{M}_\text{part}$. FreeArt3D takes the joint type as input and jointly optimizes these geometries along with the joint parameters $\mathcal{J}$, which vary depending on the joint type. For \textit{revolute} joints, $\mathcal{J}$ includes a unit rotation axis vector $\mathbf{a} \in \mathbb{R}^3$ and a pivot point $\mathbf{p} \in \mathbb{R}^3$ on the axis. The joint state $\theta_k \in \mathbb{R}$ then denotes the rotation angle (in radians) around $\mathbf{a}$. For \textit{prismatic} joints, $\mathcal{J}$ includes only the translation axis $\mathbf{a} \in \mathbb{R}^3$. The joint state $\theta_k \in \mathbb{R}$ then denotes the translation magnitude along $\mathbf{a}$. If the joint states $\theta_k$ are not available, we additionally optimize them jointly with other variables as part of the reconstruction process. 

During each optimization iteration, FreeArt3D samples an input image $\mathbf{I}_k$ and transforms the movable part according to the corresponding joint state $\theta_k$, combining it with the static body to form an articulated mesh. These image-mesh pairs are then passed to a frozen 3D diffusion model, which provides gradient signals to guide the optimization of both geometry and joint parameters.

In Section~\ref{sec:trellis}, we introduce Trellis—a pre-trained 3D diffusion model originally developed for static object generation—which serves as the supervisory signal in our framework. Section~\ref{sec:optimization} outlines our optimization pipeline for coarse geometry (in occupancy space), joint parameters $\mathcal{J}$, and articulation states $\theta_k$. To ensure convergence, Section~\ref{sec:joint_init} presents our strategies for normalizing shape scale across joint states and initializing joint parameters and geometries. These two strategies enhance the robustness of the optimization process. Finally, Section~\ref{sec:texture} details the post-processing steps used to clean the optimized occupancy grid and generate high-quality meshes with fine-grained geometry and textures.

\vspace{-0.3em}
\subsection{Preliminary: Trellis – 3D Diffusion for Static Objects}
\vspace{-0.3em}
\label{sec:trellis}

Trellis is a two-stage 3D diffusion framework designed to generate high-quality, textured 3D meshes of static objects from either a single image or a text prompt.

\noindent\textbf{Stage 1: Coarse Geometry Generation.} 
Trellis first models the coarse 3D structure using an occupancy grid of resolution $64^3$. A variational autoencoder (VAE) is trained to encode and decode this grid through a bottleneck latent:
\begin{equation}
\small
z = \text{E}_{\text{occ}}(x) \in \mathbb{R}^{16 \times 16 \times 16 \times c}, \quad \hat{x} = \text{D}_{\text{occ}}(z),
\end{equation}
where $x \in \mathbb{R}^{64 \times 64 \times 64}$ is the input occupancy grid, $\text{E}_{\text{occ}}$ and $\text{D}_{\text{occ}}$ are the encoder and decoder of the VAE respectively, $z$ is the latent representation with spatial resolution $16^3$ and feature dimension $c$, and $\hat{x}$ is the reconstructed occupancy grid. A rectified flow model $\mathcal{RF}_{\text{occ}}$ is then trained in this latent space to enable conditional generation.

\noindent\textbf{Stage 2: Detailed Geometry and Texture.}
To capture fine-grained geometry and appearance, Trellis constructs a sparse feature volume 
\[
\small
\mathcal{F} = \{ (\mathbf{x}_i, \mathbf{f}_i) \}_{i=1}^N, \quad \mathbf{x}_i \in \{0, \dots, 63\}^3, \quad \mathbf{f}_i \in \mathbb{R}^d,
\]
where $\mathcal{F}$ is a set of $N$ feature-voxel pairs, $\mathbf{x}_i$ denotes the 3D voxel coordinate of the $i$-th occupied voxel, and $\mathbf{f}_i$ is the associated $d$-dimensional feature vector extracted from multi-view DINO embeddings. This sparse volume is encoded into a sparse latent representation using a sparse encoder–decoder pair:
\begin{equation}
\small
\mathcal{Z}' = \text{E}_{\text{spa}}(\mathcal{F}) = \{ (\mathbf{x}_i, \mathbf{z}'_i) \}_{i=1}^N,\quad \mathbf{z}'_i \in \mathbb{R}^{d'}, \quad \hat{\mathcal{F}} = \text{D}_{\text{spa}}(\mathcal{Z}'),
\end{equation}
where $\text{E}_{\text{spa}}$ and $\text{D}_{\text{spa}}$ are the sparse encoder and decoder, respectively. $\mathcal{Z}'$ denotes the sparse latent feature representation, where each $\mathbf{z}'_i$ is a $d'$-dimensional embedding corresponding to voxel $\mathbf{x}_i$. The decoder reconstructs the sparse feature volume $\hat{\mathcal{F}}$ from $\mathcal{Z}'$.

A second rectified flow model, $\mathcal{RF}_{\text{spa}}$, is trained on this sparse latent space to enable conditional generation. The decoder, $\text{D}_{\text{spa}}$, then upsamples the sparse latent representation into a high-resolution sparse volume of size $256^3$. Trellis employs different decoders to support multiple outputs—for example, predicting FlexiCubes coefficients for explicit mesh generation or 3D Gaussian Splatting parameters for textured rendering.

\vspace{-0.3em}
\subsection{Optimization of Coarse Geometry and Joint}
\label{sec:optimization}
\vspace{-0.3em}

\noindent\textbf{Formulation.}  
This section focuses on optimizing the coarse geometry of both the static body $\mathcal{M}_\text{body}$ and the movable part $\mathcal{M}_\text{part}$ within the occupancy space, which corresponds to the first stage of Trellis. We also jointly optimize the joint parameters $\mathcal{J}$ as defined in Section~\ref{sec:overview}. Fine-grained geometry and texture generation are discussed in Section~\ref{sec:texture}.

To represent coarse geometry, we employ two continuous and optimizable multi-level hash grids: $\mathcal{H}_\text{body}$ and $\mathcal{H}_\text{part}$, which model the occupancy fields of the static body and movable part respectively, in the canonical (rest) frame. Each grid maps a 3D coordinate $\mathbf{c} \in \mathbb{R}^3$ to a continuous occupancy value in $[0, 1]$:
\begin{equation}
\small
\text{Occ}_\text{body}(\mathbf{c}) = \mathcal{H}_\text{body}(\mathbf{c}), \quad 
\text{Occ}_\text{part}(\mathbf{c}) = \mathcal{H}_\text{part}(\mathbf{c}).
\end{equation}

\noindent\textbf{Occupancy Grid Construction.}
At each optimization iteration, we randomly sample an input image $\mathbf{I}_k$ and its associated joint state $\theta_k$. A $64^3$ occupancy grid $x$ is then constructed by querying the values of both hash grids at the voxel coordinate $\mathbf{c}$ in the posed frame and taking the maximum occupancy value:
\begin{equation}
\small
x(\mathbf{c}) = \max\left( \text{Occ}_\text{body}(\mathbf{c}), \, \text{Occ}_\text{part}(\mathcal{T}_{\theta_k}^{-1}(\mathbf{c})) \right),
\label{equ:occupancy_grid}
\end{equation}
where $\mathcal{T}_{\theta_k}^{-1}$ denotes the inverse articulation transform, mapping voxel coordinate $\mathbf{c}$ from the posed frame back to the canonical frame based on the joint parameters $\mathcal{J}$ and joint state $\theta_k$. Specifically, for a revolute joint, the transformation is: $\mathcal{T}_{\theta_k}^{-1}(\mathbf{c}) = \mathbf{R}_{\theta_k}^{-1}(\mathbf{c} - \mathbf{p}) + \mathbf{p},$ where $\mathbf{p}$ is the pivot point and $\mathbf{R}_{\theta_k}$ is the rotation around axis $\mathbf{a}$ by angle $\theta_k$. For a prismatic joint, the transformation is: $ \mathcal{T}_{\theta_k}^{-1}(\mathbf{c}) = \mathbf{c} - \theta_k \cdot \mathbf{a}$, where $\mathbf{a}$ is the translation axis. 

\noindent\textbf{Loss Functions.}
The constructed occupancy grid $x$ is then encoded into a latent representation using the VAE encoder $\text{E}_{\text{occ}}$, i.e., $z = \text{E}_{\text{occ}}(x).$ This latent vector $z$ is paired with the input image $\mathbf{I}_k$ and passed to the frozen occupancy rectified flow model $\mathcal{RF}_{\text{occ}}$ of Trellis, which provides gradient supervision indicating what a valid 3D shape corresponding to $\mathbf{I}_k$ should be. Following the Score Distillation Sampling (SDS) procedure from DreamFusion~\cite{poole2022dreamfusion}, we perturb the latent with noise and compute the SDS loss:
\begin{equation}
\small
\nabla_\psi \mathcal{L}_{\mathrm{SDS}}(\mathcal{RF}_{\text{occ}}, \mathbf{z}) 
\triangleq \mathbb{E}_{t, \boldsymbol{\epsilon}} \left[
w(t) \cdot \left( \hat{\boldsymbol{\epsilon}}_{\mathcal{RF}_{\text{occ}}}(\mathbf{z}_t; \mathbf{I}_k, t) - \boldsymbol{\epsilon} \right) 
\cdot \frac{\partial \mathbf{z}_t}{\partial \psi}
\right],
\end{equation}
where $\psi$ denotes all optimizable parameters, including the weights of both hash grids and the joint parameters $\mathcal{J}$, $\mathbf{z} = \text{E}_{\text{occ}}(x)$ is the latent encoding of the occupancy grid from Equation~\ref{equ:occupancy_grid}, $\mathbf{z}_t$ is the noisy latent at time $t$, $\boldsymbol{\epsilon} \sim \mathcal{N}(0, \mathbf{I})$ is a sampled Gaussian noise, $\hat{\boldsymbol{\epsilon}}_{\mathcal{RF}_{\text{occ}}}(\mathbf{z}_t; \mathbf{I}_k, t)$ is the noise predicted by the rectified flow model conditioned on the image and timestep, $w(t)$ is a weight function balancing contributions across diffusion steps.

Additionally, inspired by~\cite{zhu2023hifa}, we introduce a voxel-space reconstruction loss to encourage consistency between the synthesized occupancy $x$ and the denoised prediction produced by the rectified flow model $\mathcal{RF}_{\text{occ}}$:
\begin{equation}
\small
\mathcal{L}_{\text{vox}} = \left\| \text{D}_{\text{occ}}(\hat{\mathbf{z}}_0) - x \right\|_2^2,
\end{equation}
where $\hat{\mathbf{z}}_0$ is the denoised latent derived from the predicted noise $\hat{\epsilon}_{\mathcal{RF}}(\mathbf{z}_t; \mathbf{I}_k, t)$ by the occupancy rectified flow model. This voxel-based loss improves the stability of the optimization process.

The total loss for optimizing the coarse geometry and joint parameters is given by:
\begin{equation}
\label{equ:loss}
\small
\mathcal{L}_{\text{total}} = \lambda_{\text{SDS}} \cdot \mathcal{L}_{\text{SDS}} + \lambda_{\text{vox}} \cdot \mathcal{L}_{\text{vox}},
\end{equation}
where $\lambda_{\text{SDS}}$ and $\lambda_{\text{vox}}$ are scalar weights. This optimization is repeated over randomly sampled views, progressively refining the geometry and joint parameters of the object. 

\noindent\textbf{Discussion.}
While original SDS methods utilize 2D diffusion models to guide 3D generation, FreeArt3D can be viewed as a natural extension of 3D diffusion guidance to 4D generation by treating the articulation state as an additional dimension. Unlike traditional SDS-based pipelines, which suffer from multi-view inconsistencies—such as the Janus problem—due to limited and ambiguous 2D supervision, our method avoids these issues by directly providing the full 3D shape corresponding to each joint state to the 3D diffusion model. This richer supervision signal reduces ambiguity, improves consistency, and significantly accelerates convergence, while our explicit kinematic transformations ensure structural correctness and effective disentanglement of different geometries.

\vspace{-0.3em}
\subsection{Normalization and Initialization}
\label{sec:joint_init}
\vspace{-0.3em}

\begin{figure}[t]
\includegraphics[width=\linewidth]{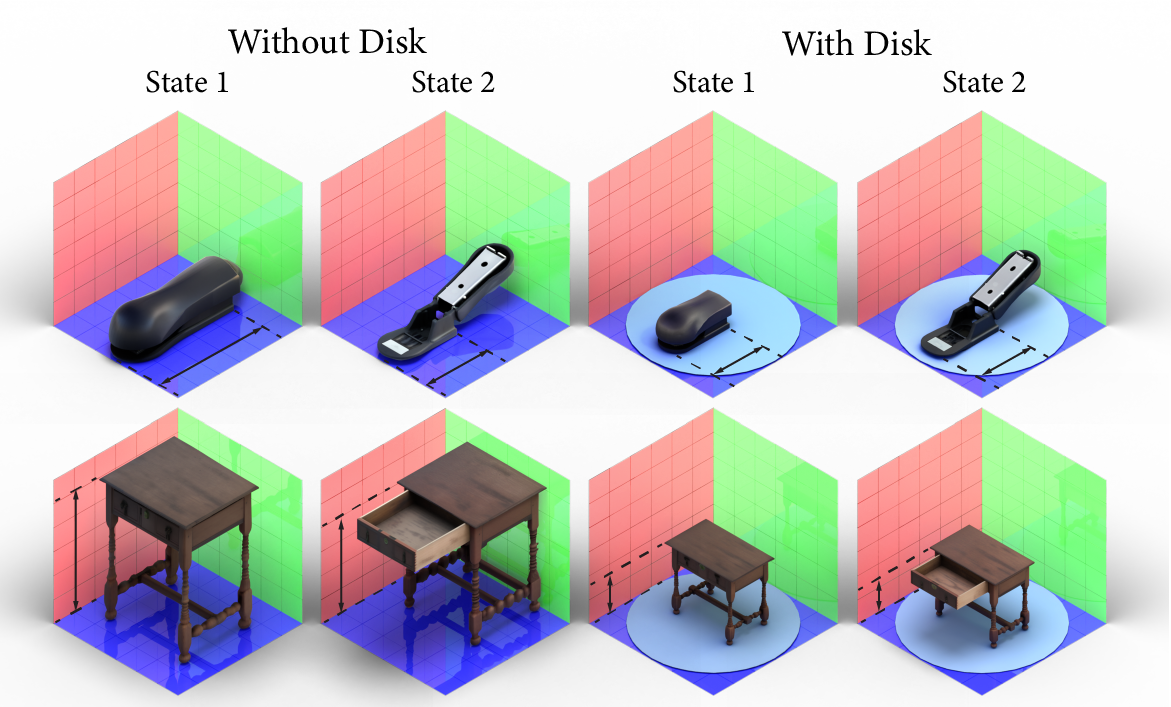}
\vspace{-2em}
\caption{\textbf{Trellis inference results across different joint states.}  Since Trellis is trained on 3D data normalized to a unit cube, each articulated component (e.g., the body of a stapler or a desk) may appear at different scales across joint states, depending on how far the movable parts (e.g., the stapler handle or desk drawer) extend. This scale inconsistency hinders optimization convergence. To address this issue, we introduce a disk beneath the object that serves as a reference to support the entire cube, ensuring consistent component scales across different states.}
\vspace{-.5em}
\label{fig:disc}
\end{figure}

\noindent\textbf{Normalization across Joint States.}
A key challenge in optimizing articulated objects lies in the significant scale variation across different joint states. This issue arises because Trellis normalizes all training shapes to fit within a unit bounding box. Consequently, the same object in different articulation states—such as a closed cabinet versus one with an open door—may result in drastically different scales for the static body part, depending on how much the movable part extends, as shown in Figure~\ref{fig:disc}. Such scale inconsistencies can severely hinder optimization, as they disrupt the geometric correspondence across joint states and may even cause optimization failure (e.g., collapsing to zero or full occupancy). 

To mitigate this, we adopt a simple yet effective strategy inspired by common practices in 3D asset creation: we add a large, fixed reference disk (resembling a carpet or floor) beneath the object in all input views. This auxiliary geometry acts as a visual anchor, providing a consistent reference for scale and spatial alignment across different joint states. By including this "carpet" in every view, the 3D diffusion model learns to interpret object scale relative to a fixed element, effectively stabilizing the predicted size of the articulated components throughout the optimization. Importantly, the carpet can be safely removed after convergence, ensuring it does not affect the final reconstruction results. This normalization strategy leads to more consistent geometric predictions across varying articulation configurations and significantly improves convergence stability.

\noindent\textbf{Initialization of Joint Parameters and Geometry.}
The initialization of both the joint parameters and the geometry plays a crucial role in ensuring robust convergence during optimization. To obtain initial estimates of the joint parameters $\mathcal{J}$ and joint states $\theta_i$ (if not provided), we employ a simple yet effective heuristic based on cross-state correspondences. Specifically, we first run full Trellis inference independently on each input image to generate textured meshes corresponding to different joint states. These meshes are then rendered into 2D images, and we apply an off-the-shelf 2D correspondence detector, such as LoFTR~\cite{sun2021loftr}, to identify pixel-level correspondences across views. The resulting 2D matches are subsequently lifted into 3D point pairs. We then filter out static point pairs and retain only those corresponding to the movable part. Given these 3D point pairs, we estimate $\mathcal{J}$ and $\theta_i$ by constructing transformations based on these parameters and minimizing the 3D distances between the corresponding points. While this procedure may yield imperfect estimates—due to factors such as inconsistencies in the generated meshes or spurious correspondences—it nonetheless provides a strong initialization for subsequent optimization, as both $\mathcal{J}$ and $\theta_i$ will be further refined jointly with the geometry.

For geometry initialization, we use the occupancy grid inferred from the image corresponding to the initial joint state as a proxy for both the static body and the movable part. This provides a reasonable geometric scaffold from which the optimization process can proceed.

\vspace{-0.3em}
\subsection{Fine-grained Geometry and Texture Generation.}
\label{sec:texture}
\vspace{-0.3em}

After optimizing the coarse geometry and articulation parameters as described in Sec.~\ref{sec:optimization}, we proceed to generate high-fidelity textured meshes for both the static and movable parts using Stage 2 of Trellis.

\input{tables/main_comparison}

\noindent\textbf{Occupancy Denoising and Cleaning.}  
We first construct the merged occupancy grid $x_{\text{max}}$ corresponding to the largest joint state $\theta_{\text{max}}$, using the optimized hash grids and Equation~\ref{equ:occupancy_grid}. To improve voxel quality, we inject small Gaussian noise into $x_{\text{max}}$ and denoise it via a single forward pass through the occupancy rectified flow model $\mathcal{RF}_{\text{occ}}$, yielding a cleaner occupancy grid $\tilde{x}$. We then remove auxiliary geometry—specifically the "carpet plane"—by detecting and discarding voxels near the lowest $z$-values in $\tilde{x}$. Finally, we filter outlier voxels based on spatial isolation and size thresholds to produce a clean voxel grid $\bar{x}$.

\noindent\textbf{Sparse Latent Construction and Decoding.}  
From the cleaned occupancy grid $\bar{x}$, we extract the set of occupied voxel coordinates, and construct an initial sparse latent feature volume by sampling Gaussian noise. This initial sparse latent is then denoised using the second-stage flow model $\mathcal{RF}_{\text{spa}}$ of Trellis. We then decode the denoised latent using the sparse decoder $\text{D}_{\text{spa}}$ to produce a sparse feature volume $\hat{\mathcal{F}}$, containing both FlexiCubes coefficients for mesh extraction and Gaussian Splatting parameters for texture synthesis.

\noindent\textbf{Mesh Extraction and Texture Baking.}  
To recover part-specific geometries, we partition the decoded sparse volume $\hat{\mathcal{F}}$ into two subsets corresponding to the static body and the movable part. Specifically, we examine the transformed occupancy grids of the two parts, assigning each cell to the part with higher occupancy. For each part, we extract a detailed mesh from the associated FlexiCubes representation. We then bake the surface textures onto each mesh using the decoded Gaussian Splatting parameters of the complete shape. Finally, we combine the resulting textured meshes with the optimized joint parameters $\mathcal{J}$ to form the complete articulated object with fine-grained geometry and realistic appearance. This concludes the FreeArt3D pipeline.

%% file: tables/main_comparison.tex
\begin{table*}[t]
  \centering
  \scriptsize
  \setlength{\tabcolsep}{3.5pt}
  \caption{\textbf{Quantitative Comparison between ParisArticulate-Anything~\cite{le2024articulate}, Singapo~\cite{liu2024singapo}, URDFormer~\cite{chen2024urdformer}, PARIS~\cite{liu2023paris} and Ours.} ``--'' indicates that the baseline method fails to handle the given category or the metric is not available. ``Average-5'' refers to the average performance over the five categories that all methods can handle. ``Average-12'' refers to the average performance over the twelve categories that Articulate-Anything, PARIS and our method can handle.
}
\vspace{-1.5em}
    \begin{tabular}{c|c|cccccccccccc|c|c}
    \toprule
          &       & Box   & Dishwasher & Laptop & Lighter & Microwave & Oven  & Refrigerator & Safe  & Stapler & StorageFurniture & Table & WashingMachine & Average-5 & Average-12 \\
    \midrule
    \multirow{5}[2]{*}{Ours} & F-Score & \textbf{0.915} & \textbf{0.929} & \textbf{0.845} & \textbf{0.849} & \textbf{0.891} & \textbf{0.869} & \textbf{0.912} & \textbf{0.915} & \textbf{0.922} & \textbf{0.873} & \textbf{0.950} & 0.834 & \textbf{0.883} & \textbf{0.891} \\
          & CD    & \textbf{0.023} & \textbf{0.021} & \textbf{0.037} & \textbf{0.028} & \textbf{0.026} & \textbf{0.026} & \textbf{0.024} & \textbf{0.021} & \textbf{0.019} & 0.026 & \textbf{0.018} & 0.031 & \textbf{0.026} & \textbf{0.025} \\
          & Clip-Sim & \textbf{0.883} & \textbf{0.879} & \textbf{0.871} & \textbf{0.839} & \textbf{0.885} & \textbf{0.877} & \textbf{0.885} & \textbf{0.883} & \textbf{0.881} & \textbf{0.897} & \textbf{0.922} & 0.872 & \textbf{0.882} & \textbf{0.881} \\
          & Joint-Axis-Err & \textbf{0.021} & \textbf{0.028} & \textbf{0.143} & \textbf{0.040} & \textbf{0.021} & \textbf{0.170} & 0.151 & \textbf{0.023} & \textbf{0.039} & \textbf{0.039} & 0.535 & 0.654 & \textbf{0.208} & \textbf{0.159} \\
          & Joint-Pivot-Err & \textbf{0.140} & \textbf{0.075} & \textbf{0.071} & \textbf{0.138} & 0.150 & \textbf{0.094} & \textbf{0.029} & \textbf{0.040} & \textbf{0.305} & - & - & 0.171 & \textbf{0.092} & \textbf{0.121} \\
    \midrule
    \multirow{5}[2]{*}{ArtAnything} & F-Score & 0.642 & 0.761 & 0.693 & 0.760 & 0.781 & 0.808 & 0.824 & 0.719 & 0.754 & 0.847 & 0.748 & \textbf{0.848} & 0.817 & 0.769 \\
          & CD    & 0.062 & 0.052 & 0.056 & 0.041 & 0.038 & 0.035 & 0.035 & 0.049 & 0.049 & 0.029 & 0.042 & \textbf{0.030} & 0.036 & 0.043 \\
          & Clip-Sim & 0.837 & 0.841 & 0.805 & 0.792 & 0.855 & 0.858 & 0.865 & 0.843 & 0.828 & 0.938 & 0.857 & \textbf{0.877} & 0.876 & 0.851 \\
          & Joint-Axis-Err & 0.564 & 0.651 & 0.227 & 0.264 & 0.487 & 0.958 & 0.164 & 0.372 & 1.119 & 0.182 & 0.287 & 0.681 & 0.527 & 0.499 \\
         & Joint-Pivot-Err & 0.265 & 0.179 & 0.227 & 0.141 & \textbf{0.057} & 0.122 & 0.074 & 0.087 & 0.586 & - & - & 0.168 & 0.136 & 0.191 \\
    \midrule
    \multirow{5}[2]{*}{Singapo} & F-Score & -     & 0.848 & -     & -     & 0.805 & 0.789 & 0.869 & -     & -     & 0.881 & 0.859 & 0.774 & 0.832 & - \\
          & CD    & -     & 0.030 & -     & -     & 0.036 & 0.035 & 0.029 & -     & -     & \textbf{0.023} & 0.026 & 0.036 & 0.031 & - \\
          & Clip-Sim & -     & 0.873 & -     & -     & 0.856 & 0.856 & 0.860 & -     & -     & 0.880 & 0.846 & 0.827 & 0.859 & - \\
          & Joint-Axis-Err & -     & 0.285 & -     & -     & 0.145 & 0.340 & \textbf{0.116} & -     & -     & 0.056 & \textbf{0.081} & \textbf{0.439} & 0.247 & - \\
          & Joint-Pivot-Err & - & 0.095 & - & - & 0.077 & 0.147 & 0.034 & - & - & - & - & 0.101 & 0.094 & - \\
     \midrule
     \multirow{4}[2]{*}{URDFormer} & F-Score & -     & 0.678 & -     & -     & -     & 0.703 & 0.536 & -     & -     & 0.740 & -     & 0.737 & 0.679 & - \\
          & CD    & -     & 0.059 & -     & -     & -     & 0.047 & 0.076 & -     & -     & 0.044 & -     & 0.044 & 0.054 & - \\
          & Clip-Sim & -     & 0.846 & -     & -     & -     & 0.800 & 0.826 & -     & -     & 0.810 & -     & 0.805 & 0.817 & - \\
          & Joint-Axis-Err & -     & 1.301 & -     & -     & -     & 1.268 & 1.331 & -     & -     & 1.084 & -     & 1.306 & 1.258 & - \\
          & Joint-Pivot-Err & - & 0.360 & - & - & - & 0.414 & 0.293 & - & - & - & - & 0.312 & 0.344 & - \\
    \midrule
    \multirow{5}[2]{*}{PARIS} & F-Score & 0.723 & 0.729 & 0.834 & 0.897 & 0.762 & 0.810 & 0.740 & 0.760 & 0.836 & 0.806 & 0.949 & 0.797 & 0.776 & 0.804 \\
          & CD & 0.047 & 0.047 & 0.043 & 0.022 & 0.038 & 0.033 & 0.036 & 0.037 & 0.039 & 0.030 & 0.016 & 0.030 & 0.035 & 0.035 \\
          & Clip-Sim & 0.755 & 0.790 & 0.762 & 0.785 & 0.776 & 0.771 & 0.807 & 0.775 & 0.779 & 0.751 & 0.778 & 0.798 & 0.783 & 0.777 \\
          & Joint-Axis-Err & 1.090 & 1.334 & 1.014 & 0.490 & 1.413 & 1.283 & 1.212 & 1.464 & 0.421 & 1.334 & 1.537 & 1.072 & 1.247 & 1.139 \\
          & Joint-Pivot-Err & 0.166 & 0.169 & 0.203 & 0.356 & 0.385 & 0.138 & 0.074 & 0.216 & 0.349 & - & - & 0.364 & 0.186 & 0.242  \\
    \bottomrule
    \end{tabular}%
    \vspace{-0.7em}
  \label{tab:main_comparison}%
\end{table*}%

%% file: sections/experiment.tex
\begin{figure*}
    \centering
    \includegraphics[width=\linewidth]{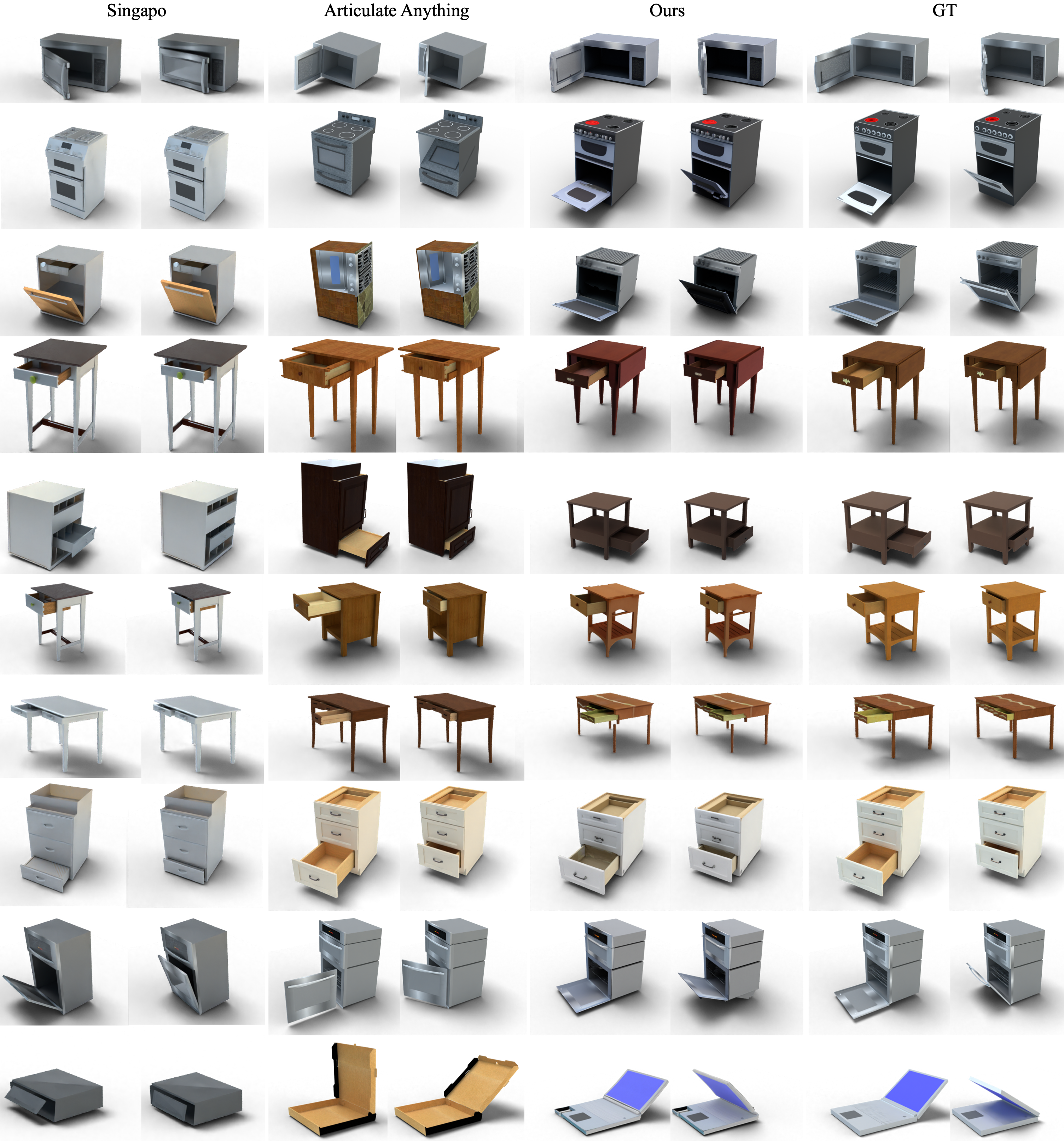}
    \caption{\textbf{Comparison between Singapo~\cite{liu2024singapo}, Articulate-Anything~\cite{le2024articulate}, and Ours.} Unlike baseline methods that rely on part retrieval and fail to reconstruct detailed geometry and textures, our method generates meshes that closely match the input images and successfully recover fine-grained geometric details, realistic textures, and accurate articulation structures. } 
    \label{fig:main-comp}
\end{figure*}
 
\section{Experiment} 
\subsection{Implementation Details and Evaluation Setup}

\noindent\textbf{Implementation Details.}  
For each shape, we use $K = 6$ input images $\mathbf{I}_k$, each corresponding to a different joint state $\theta_k$, sampled between the rest and maximum configurations. Our method only requires images captured at different articulation states, not necessarily from different camera views, although view variation is allowed, as shown in the teaser. In our experiments, however, the camera pose is fixed to a single side–upper view. We assume the joint states $\theta_k$ are unknown and jointly optimize them along with the joint parameters $\mathcal{J}$. Optimization is performed over 3000 iterations per shape using a learning rate of 0.01 and the Adam optimizer~\cite{kingma2014adam}. The loss weights $\lambda_{\text{SDS}}$ and $\lambda_{\text{VOX}}$ are set to 0.1 and 1.0, respectively. SDS noise is sampled from the range $[0.5, 0.8]$ before scaling. To estimate the initial joint parameters, we render 18 images—3 different views per state—using Blender, and select 36 image pairs in total to build correspondences and estimate the initialization. We adopt TinyCudaNN~\cite{tiny-cuda-nn} for hash grid implementation. To clean the occupancy voxels, we inject small noise of 0.5 before scaling. The total runtime for processing a single shape is approximately 10 minutes.

\noindent\textbf{Evaluation Dataset and Metric}  
We evaluate on the PartNet-Mobility dataset~\cite{xiang2020sapien}, focusing on 12 categories: box, dishwasher, laptop, lighter, microwave, oven, refrigerator, safe, stapler, storage furniture, table, and washing machine. For each category, we randomly sample 12 shapes, resulting in 144 test shapes in total. Given a generated geometry pair (body and movable parts) and predicted joint parameters, we compute five metrics: Chamfer Distance (CD)~\cite{fan2017point}, F-Score~\cite{wang2018pixel2mesh}, CLIP similarity, joint axis direction error, and joint pivot error. For each shape, we sample 6 joint states and generate corresponding meshes using the estimated joint parameters. We compute the metrics for each state and report their average.

Before evaluation, we align the generated mesh and ground-truth mesh using the method from~\cite{liu2024one}. CD and F-Score measure geometric similarity—CD is computed by sampling 100k surface points, while F-Score uses a 0.05 threshold. For appearance, we render 5 views of both predicted and ground-truth meshes and compute the average CLIP similarity~\cite{radford2021learning} using ViT-L/14@336px. Following ArticulateAnything~\cite{le2024articulate}, we report the joint axis direction error as the angle between the predicted axis $\mathbf{a}_p$ and the ground-truth axis $\mathbf{a}_g$ for both revolute and prismatic joints:
\begin{equation}
\small
e_{\text{axis}} = \min \left( 
\arccos\left( \frac{\mathbf{a}_p \cdot \mathbf{a}_g}{\|\mathbf{a}_p\|_2 \|\mathbf{a}_g\|_2} \right), 
\arccos\left( \frac{-\mathbf{a}_p \cdot \mathbf{a}_g}{\|\mathbf{a}_p\|_2 \|\mathbf{a}_g\|_2} \right)
\right),
\end{equation}
where $e_{\text{axis}} \in [0, \pi]$. For revolute joints, the joint pivot error is defined as the shortest distance between the predicted and ground-truth joint axes:
\begin{equation}
\small
e_{\text{pivot}} =
\frac{|\mathbf{p} \cdot (\mathbf{a}_p \times \mathbf{a}_g)|}{|\mathbf{a}_p \times \mathbf{a}_g|},
\end{equation}
where $\mathbf{p} = \mathbf{x}_p - \mathbf{x}_g$ denotes the vector difference between the predicted and ground-truth pivot points. Note that CD and F-Score also implicitly reflect joint prediction accuracy, as they are evaluated per joint state.

\vspace{-0.3em}
\subsection{Comparison with Baselines}
\vspace{-0.3em}

\noindent \textbf{Baselines}
We compare our method against four state-of-the-art approaches for modeling articulated objects: Articulate-Anything~\cite{le2024articulate}, Singapo~\cite{liu2024singapo}, URDFormer~\cite{chen2024urdformer}, and PARIS~\cite{liu2023paris}. The first three methods take a single-view image as input, whereas PARIS reconstructs from multi-view images. Both Articulate-Anything and Singapo assemble articulated objects by retrieving parts from the PartNet-Mobility dataset~\cite{xiang2020sapien} and applying the textures of the retrieved parts. URDFormer, on the other hand, generates articulated objects by predicting bounding boxes for parts, then scaling and composing template part meshes. It directly projects the input image onto the mesh surface to serve as the texture. PARIS reconstructs the object’s shape and appearance by jointly optimizing two neural fields—one for the static parts and one for the movable parts. For a fair comparison, we remove the test shapes from the retrieval database for both Articulate-Anything and Singapo. We also use a relatively sparse set of 24 views for PARIS. Among the 12 categories, Articulate-Anything and PARIS supports all, Singapo supports 7 out of 12, and URDFormer supports only 5 out of 12.
\input{tables/runtime}

\noindent \textbf{Results}
As shown in Figure~\ref{fig:main-comp}, our method generates high-quality textured meshes that closely match the input images and, consequently, closely resemble the ground-truth meshes. In contrast, both Singapo and Articulate-Anything do not generate new geometry or textures for the input images; instead, they retrieve parts from a database—a process fundamentally constrained by the database's scale. Due to the scarcity of 3D articulated objects, these methods often produce results that capture only the coarse global shape while failing to reproduce fine-grained geometric details, colors, and textures. Such limitations significantly hinder their applicability in downstream tasks such as digital twins for robotics, which require accurate reconstructions of real-world environments. An interesting observation is that all retrieval-based baseline methods successfully retrieved the exact same or highly similar object from the database (Row 8), whereas our method—despite not relying on retrieval—still produces results of comparable quality. This highlights the robustness and generalization capability of our approach. Furthermore, both Singapo and Articulate-Anything frequently fail to predict accurate joint axes, as illustrated in Rows 2, 3, and 9.

Quantitative results are shown in Table~\ref{tab:main_comparison}, where our method significantly outperforms all baseline methods across all five metrics. The retrieval-based methods, Articulate-Anything and Singapo, perform considerably better than the template-mesh-based method URDFormer and optimized-based PARIS, yet still fall far short of our generative model. Notably, while baseline methods are limited to a small number of predefined categories, FreeArt3D demonstrates strong generalizability and can handle a much broader range of categories. We also report the runtimes of all methods in Table~\ref{tab:runtime}.

\begin{figure*}[t]
    \centering
    \centering
    \includegraphics[width=\linewidth]{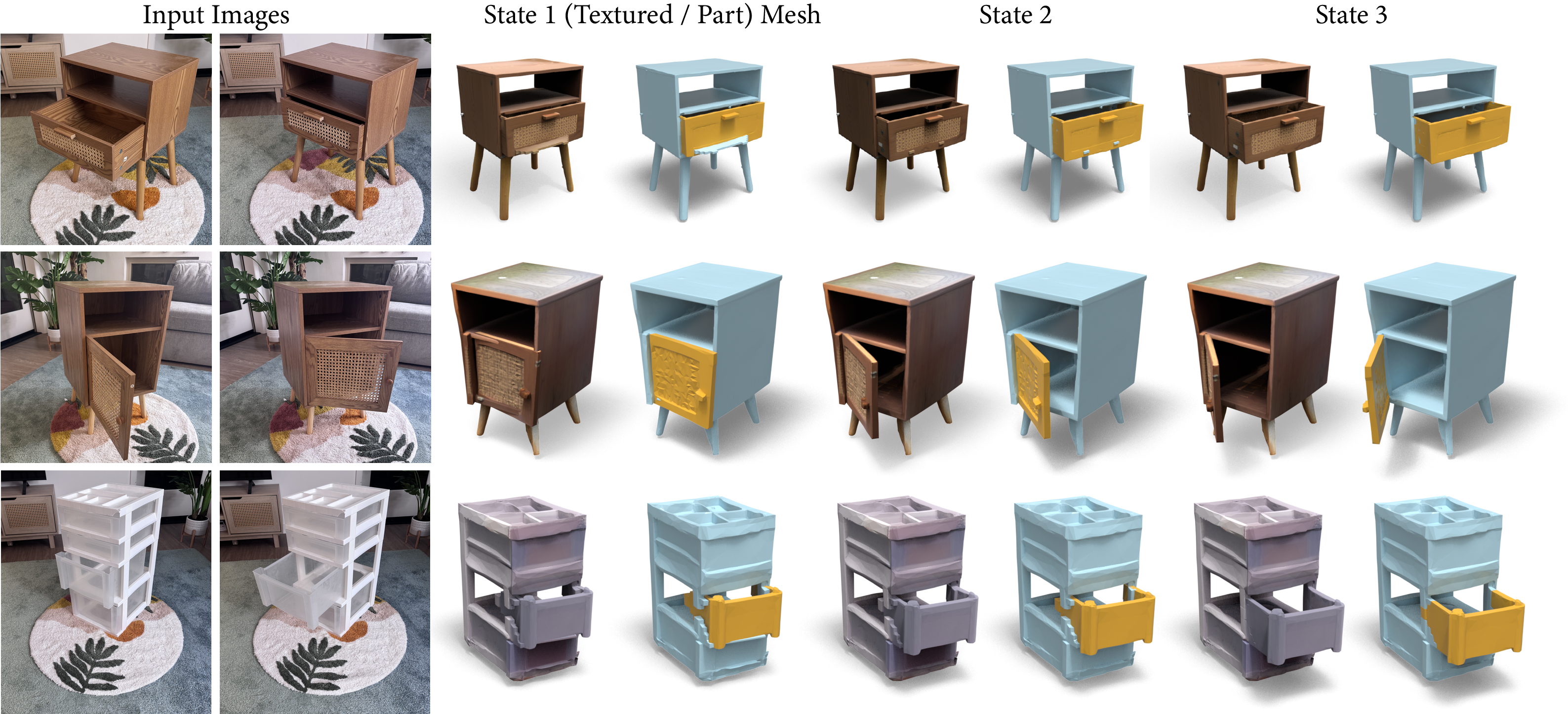}
    \vspace{-2em}
    \caption{\textbf{Real-World Demo}. For each shape, we capture six images of the object in different joint states. FreeArt3D effectively leverages these casually captured, unposed images to generate high-quality articulated objects with sharp geometry and realistic textures.} 
    \vspace{-1em}
    \label{fig:real-world}
\end{figure*}

\begin{figure}
    \centering
    \includegraphics[width=\linewidth]{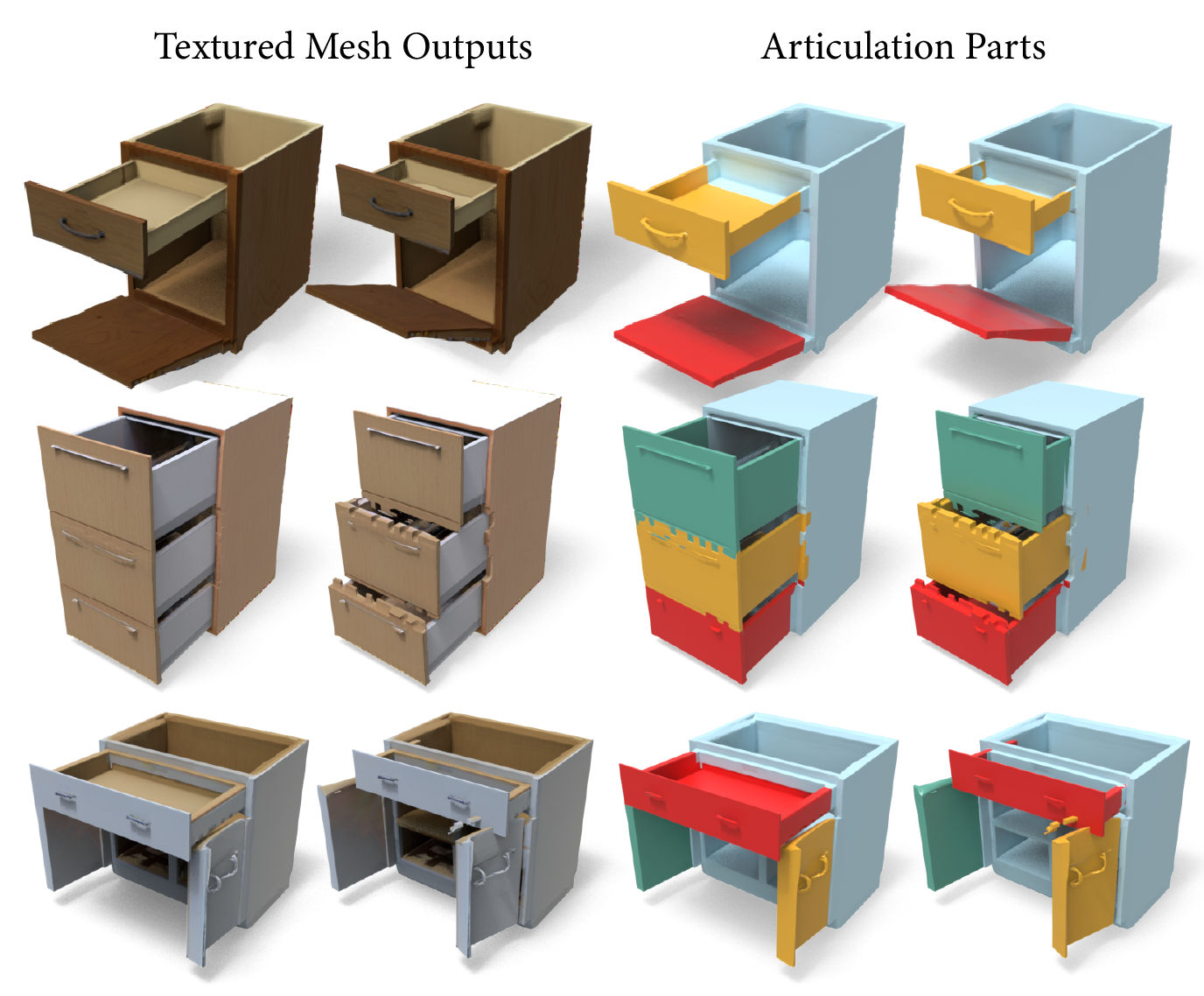}
    \vspace{-2.5em}
    \caption{\textbf{Generation Results of Multiple Parts and Joints.} Our method can be easily extended to support the generation of multiple articulated parts and joints, enabling flexible configuration of all components in the generated objects.}
    \label{fig:multipart}
\end{figure}

\subsection{Application and Extension}
\noindent\textbf{Multi-Joint Extension.} In our method and previous experiments, we primarily focus on articulated objects with a single prismatic or revolute joint for simplicity. However, extending our approach to more complex objects with multiple joints and mixed joint types-while still assuming a single–degree-of-freedom kinematic chain-is straightforward. To enable such an extension, several adaptations are necessary: (a) using more input images to capture the motion of each joint; (b) employing a separate hash grid to represent the occupancy of each part; and (c) jointly optimizing the geometry of all parts along with their corresponding joint parameters. As shown in Figure~\ref{fig:multipart}, our method supports multi-joint generation, enabling the creation of complete URDFs, which facilitates downstream applications such as robotic simulation and digital twins.

\noindent\textbf{Real-World Reconstruction.} Beyond standard synthetic benchmarks, we also evaluate our method on real-world captured images. For each object, we capture six images using an iPhone, with the object in different joint states. Note that our method does not require camera poses as input, nor does it depend on dense-view images or assume full object coverage. As a result, the required sparse-view images are easy to obtain. However, we do require either a physical carpet placed beneath the object or a virtual disc added to the captured photo for normalization. Figure~\ref{fig:real-world} shows example inputs and results, demonstrating that our method can potentially support a wide range of real-world applications.

\subsection{Ablation Study and Analysis}
We report the quantitative results of our ablation study in Table~\ref{tab:ablation}.

\input{tables/ablation}

\noindent\textbf{Loss Functions.} Our coarse-geometry optimization primarily relies on a combination of the SDS loss and the voxel loss. Removing the SDS loss (row (a)) and relying solely on the voxel-space L2 loss leads to a significant drop in performance, highlighting the necessity of the SDS loss. In contrast, the voxel loss (row (b)) serves as a complementary term that helps stabilize training and further improve performance.

\noindent\textbf{Disk Normalization.} We introduce an auxiliary disk beneath the object to promote consistent scaling of the articulated components across different joint states, as discussed in Section~\ref{sec:joint_init}. The necessity of this normalization is evidenced in row (c); removing this technique results in a significant drop in performance.

Some may be concerned that adding a disk could affect Trellis’s generation capability. To verify this, we evaluated Trellis on renderings of 144 3D objects with and without disks, comparing the Chamfer Distance (CD) between the generated static 3D models and the ground truth. The results show that the average CD is 0.018 with a disk and 0.024 without a disk. Interestingly, the results with disks performed slightly better, providing no evidence of capacity degradation or bias introduced by the disk. As alternative approaches for scale normalization, one could explore 3D point correspondences or fine-tuning Trellis with other normalization schemes.

\noindent\textbf{Joint Initialization.} In Section~\ref{sec:joint_init}, we propose a preprocessing step that estimates the joint axis to serve as the initialization. In row (d), we remove this joint estimation module and instead use a random initialization, allowing the joint parameters to be optimized jointly with the occupancy volumes. We perform three trials and observe a significant performance drop, indicating that without a good initialization, relying solely on SDS and voxel losses is insufficient for accurately recovering the joint axis, as reflected in both geometric and joint metrics. In row (e), we retain the estimation module but fix the estimated joint parameters during coarse geometry optimization. This results in a slight performance drop, suggesting that our optimization process is capable of refining the initial estimated joint paramters further.

\noindent\textbf{Hash Grid.} In our preliminary exploration, we find that representing occupancy using a discrete $64^3$ voxel grid can introduce instability during optimization. In contrast, a continuous multi-level hash grid offers a smoother and more flexible representation, leading to improved stability and effectiveness in optimization. In row (f), we compare the performance of discrete voxel grids and continuous multi-level hash grids and observe a notable performance gap.

\begin{figure}
    \centering
    \includegraphics[width=\linewidth]{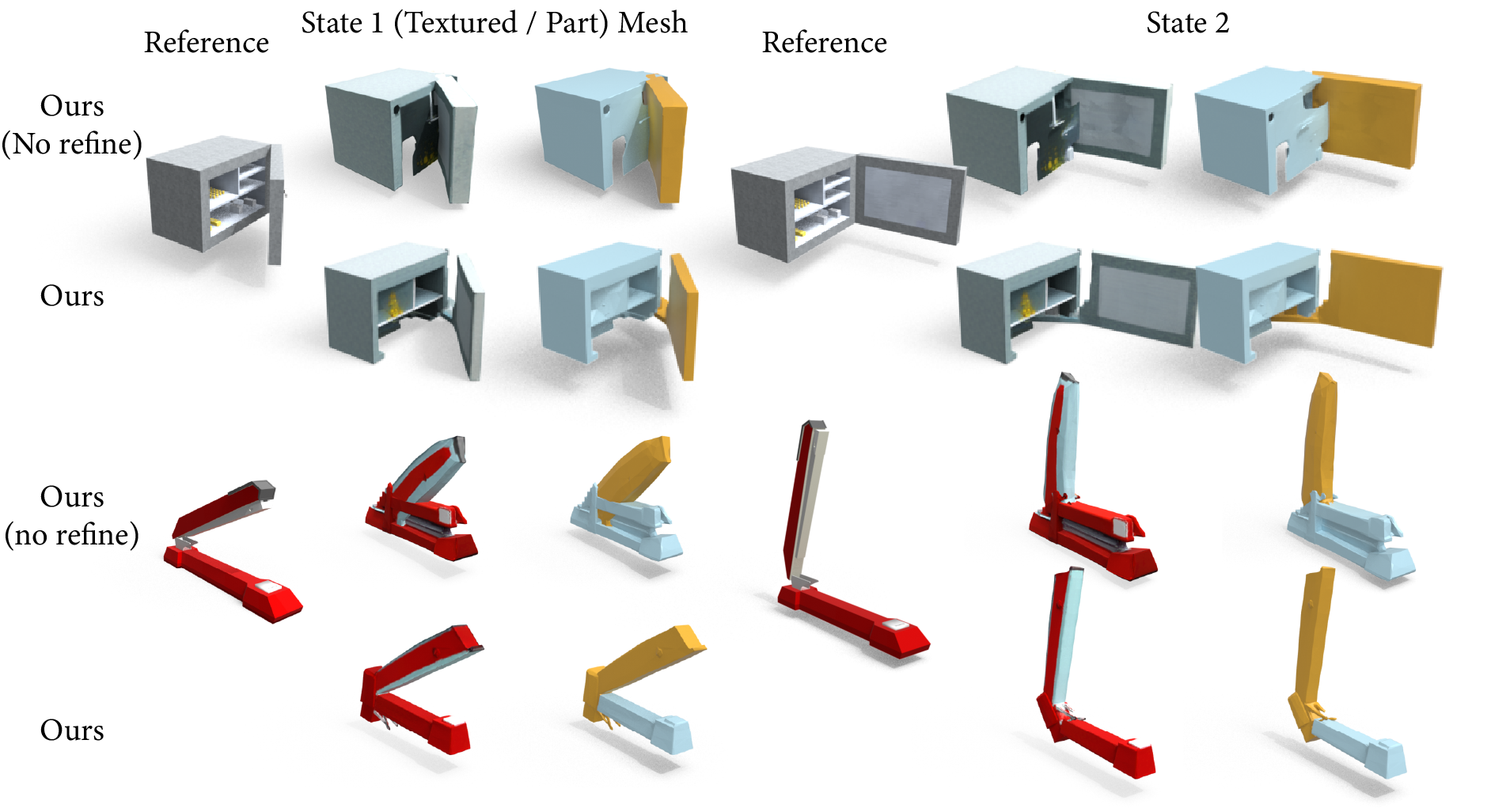}
    \vspace{-2em}
    \caption{\textbf{Ablation Study on Occupancy Refinement After Coarse Geometry Optimization.}}
    \label{fig:refinement}
\end{figure}

\noindent\textbf{Voxel Refinement.} In Sec.~\ref{sec:texture}, we introduce several techniques to refine the optimized occupancy voxels before passing them to the second-stage diffusion model for fine-grained geometry and texture generation. As shown in Row (g), this step further enhances performance. See Figure~\ref{fig:refinement} for qualitative examples.

\noindent\textbf{Number of Input Views.} In our default setup, we use 6 sparse input views, which already yield robust and satisfactory results. In Row (h) and Row (i), we demonstrate that incorporating more input views (e.g., 21 views) can further improve performance slightly, while only using the minimum views still performs reasonably well.

\noindent\textbf{Benefit of Trellis.} Our method leverages the pre-trained 3D diffusion model Trellis~\cite{xiang2024structured}. However, its success cannot be solely attributed to Trellis. In previous ablation studies, we have already demonstrated the importance of disk normalization and joint initialization. To isolate Trellis’s contribution, we modified the baseline Singapo by replacing its part retrieval module with Trellis-based part generation, and the results are presented in Table~\ref{tab:trellis_comparison}. Note that we used ground-truth, separately rendered part images as input, whereas in practice only full-object images containing all parts are available. Because of this unrealistic input setting, Singapo’s CLIP similarity is slightly better than ours. Nevertheless, our method still outperforms it on other metrics, demonstrating its overall superiority. These results indicate that directly enhancing baselines with a pre-trained 3D diffusion model is non-trivial.

\input{tables/enhanced_trellis}

\input{tables/failure_analysis}

\subsection{Discussion on Robustness and Failure Cases}
We observe three common failure cases of our method: (1) incorrect joint axis direction (angle error $> 20^{\circ}$), (2) inaccurate pivot point (distance error $> 0.1$), and (3) failed part segmentation—producing only the movable or the static part. To evaluate the robustness of our method and analyze these failures, we report the failure rates of each error type and various ablated versions in Table~\ref{tab:failure_analysis}. 

Overall, our method is relatively robust. With only 6 input images, the error rate for each type is around $10\%$, and the method achieves a 77.08\% overall success rate (meeting all three criteria). We also find that robustness strongly depends on scale normalization and the quality of joint initialization. Without the proposed disk normalization, the success rate drops significantly. When using fewer input images (e.g., 2 images), the method still works in many cases, though the success rate decreases substantially, mainly due to the difficulty of initial joint estimation. Although good joint initialization is important, the joint optimization process over both geometry and joint parameters remains essential, since it can refine imperfect initializations. Remarkably, the method sometimes succeeds even with random joint initialization.

%% file: tables/runtime.tex
\begin{table}[t]
  \centering
  \scriptsize
  \setlength{\tabcolsep}{8pt}
  \caption{\textbf{Runtime Comparison.} Runtime (seconds) of our method and baseline methods. Measured on a single NVIDIA H100 GPU.}
  \vspace{-1.5em}
  \begin{tabular}{ccccc}
    \toprule
    Ours & ArtAnything & Singapo & URDFormer & PARIS \\
    \midrule
     606s & 200s & 19s & 27s & 494s \\
    \bottomrule
  \end{tabular}
  \label{tab:runtime}
\end{table}

%% file: tables/ablation.tex
\begin{table}[t]
  \scriptsize
  \setlength{\tabcolsep}{2pt}
  \caption{\textbf{Ablation Study.} Evaluated across all categories.}
  \vspace{-1.5em}
    \begin{tabular}{c|c|ccccc}
    \toprule
    Index & Version & F-Score$\uparrow$ & CD $\downarrow$ & CLIP-Sim $\uparrow$ & Axis-Err$\downarrow$ & Pivot-Err$\downarrow$ \\
    \midrule
    a     & no SDS loss & 0.610 & 0.067 & 0.808 & 0.555 & 0.152 \\
    b     & no voxel loss & 0.873 & 0.028 & 0.880 & 0.172 & 0.138 \\
    c     & no disk normalization & 0.749 & 0.048 & 0.854 & 0.730 & 0.244 \\
    d     & rand joint initialization & 0.753 & 0.049 & 0.849 & 1.024 & 0.253 \\
    e     & fixed joint after initialization & 0.864 & 0.029 & 0.876 & 0.165 & 0.138 \\
    f     & no hashgrid & 0.848 & 0.032 & 0.872 & 0.157 & 0.180 \\
    g     & no voxel refinement & 0.875 & 0.027 & 0.858 & 0.181 & 0.111 \\
    h     & more input (6 $\rightarrow$ 21) & 0.903 & 0.023 & 0.883 & 0.126 & 0.135 \\
    i     &  minimum input (6 $\rightarrow$ 2) &  0.841 &  0.035 &  0.870 &  0.525 & 0.227 \\
    j     & full  & 0.892 & 0.025 & 0.881 & 0.155 & 0.121 \\
    \bottomrule
    \end{tabular}%
  \label{tab:ablation}%
\end{table}%

%% file: tables/enhanced_trellis.tex
\begin{table}[t]
  \centering
  \scriptsize
  \caption{\textbf{Comparison with Trellis-Enhanced Singapo. Evaluated across all categories supported by Singapo.}}
  \vspace{-1.5em}
  \begin{tabular}{c|ccccc}
    \toprule
    Method & F-Score$\uparrow$ & CD$\downarrow$ & CLIP-Sim$\uparrow$ & Joint-Axis-Err$\downarrow$ & Joint-Pivot-Err$\downarrow$ \\
    \midrule
    Original Singapo    & 0.832 & 0.031 & 0.859 & 0.247 & 0.094 \\
    Singapo w/ Trellis  & 0.834 & 0.031 & \textbf{0.899} & 0.255 & 0.109 \\
    Ours                & \textbf{0.883} & \textbf{0.026} & 0.882 & \textbf{0.208} & \textbf{0.092} \\
    \bottomrule
  \end{tabular}
  \vspace{-1em}
  \label{tab:trellis_comparison}
\end{table}

%% file: tables/failure_analysis.tex
\begin{table}[t] 
  \centering 
  \scriptsize
  \setlength{\tabcolsep}{3pt}
  \caption{\textbf{Failure Case Analysis. The percentages indicate the proportion of failure cases, evaluated across all categories.}}
  \vspace{-1.5em}
  \begin{tabular}{c|ccc}
    \toprule
    Failure Type & Axis Direction$\downarrow$ & Pivot Point$\downarrow$ & Segmentation$\downarrow$ \\
    \midrule
    Ours (w/o. disk)  & 45.14\% & 43.75\% & 41.67\% \\
    Ours (2-state)      & 32.64\% & 47.92\% & 11.81\% \\
    Ours (6-state)      &  \textbf{9.03\%} & \textbf{11.11\%} &  \textbf{9.03\%} \\
    \bottomrule
  \end{tabular}
  \label{tab:failure_analysis}
\end{table}

%% file: sections/conclusion.tex
\section{Conclusion and Limitation} 

We introduce a novel training-free pipeline for modeling articulated objects, which effectively circumvent the challenges posed by the scarcity of articulated object data. Unlike previous methods that rely on part retrieval and template meshes, our generative model enables the production of high-fidelity textured meshes that closely adhere to the input prompts. Looking ahead, it would be interesting to explore ways to further accelerate the optimization process and improve its robustness—for example, by developing more elegant solutions to normalization issues.